\documentclass{article}
\usepackage{ijcai26}

\usepackage{times}
\usepackage{soul}
\usepackage{url}
\usepackage[hidelinks]{hyperref}
\usepackage[utf8]{inputenc}
\usepackage[small]{caption}
\usepackage{graphicx}
\usepackage{amsmath}
\usepackage{amsthm}
\usepackage{booktabs}
\usepackage{algorithm}
\usepackage{algorithmic}
\usepackage[switch]{lineno}

\usepackage{booktabs}
\usepackage{amssymb}
\usepackage{caption}
\usepackage{multirow}
\usepackage{makecell}
\usepackage{graphicx}
\usepackage{tabularx}
\usepackage{array}
\usepackage{booktabs}
\usepackage{makecell}
\usepackage{cuted}
\usepackage{bibunits}

\title{ALIGN : Advanced Query Initialization with LiDAR-Image Guidance for Occlusion-Robust 3D Object Detection}

\author{
Janghyun Baek$^1$
\and
Mincheol Jang$^1$\and
Seokha Moon$^1$\and
Jinkyu Kim$^1{^*} $\and
Seungjoon Lee$^2$\\
\affiliations
$^1$Computer Scienece and Engineering, Korea University\\
$^2$CTO Devision, LG Innotek\\
\emails
\{2022020909, m1ncheoree, shmoon96, jinkyukim\}@korea.ac.kr,
lsj931228@lginnotek.com
}

\begin{document}

\maketitle
\begin{abstract}
\label{sec:abstract}
Recent query-based 3D object detection methods using camera and LiDAR inputs have shown strong performance, but existing query initialization strategies, including random and BEV heatmap-based sampling, still suffer from fundamental limitations.
Random sampling uniformly distributes queries across the scene without object awareness, leading to inefficient focus and slow convergence.
Heatmap-based sampling selects salient BEV regions but often overlooks occluded or small objects whose features are weak or invisible.
To address this limitation, we propose \textbf{ALIGN} (\textbf{A}dvanced query initialization with \textbf{L}iDAR and \textbf{I}mage \textbf{G}uida\textbf{N}ce), a novel approach for occlusion-robust, object-aware query initialization. 
Our approach aligns 3D LiDAR geometry with 2D image semantics to estimate reliable object centers and adaptively place queries around them. In addition, it maintains a controlled budget of background queries to preserve global coverage, producing a tightly coupled routine that remains effective in both sparse and crowded scenes.
Extensive experiments on the nuScenes benchmark show that ALIGN consistently improves performance across multiple state-of-the-art detectors, achieving gains of up to +0.9 mAP and +1.2 NDS, particularly under occlusion and dense traffic.
Our code will be publicly available upon publication.
\end{abstract}
\section{Introduction}
\label{sec:intro}
\begin{figure}[t]
    \centering
    \includegraphics[width=0.98\columnwidth]{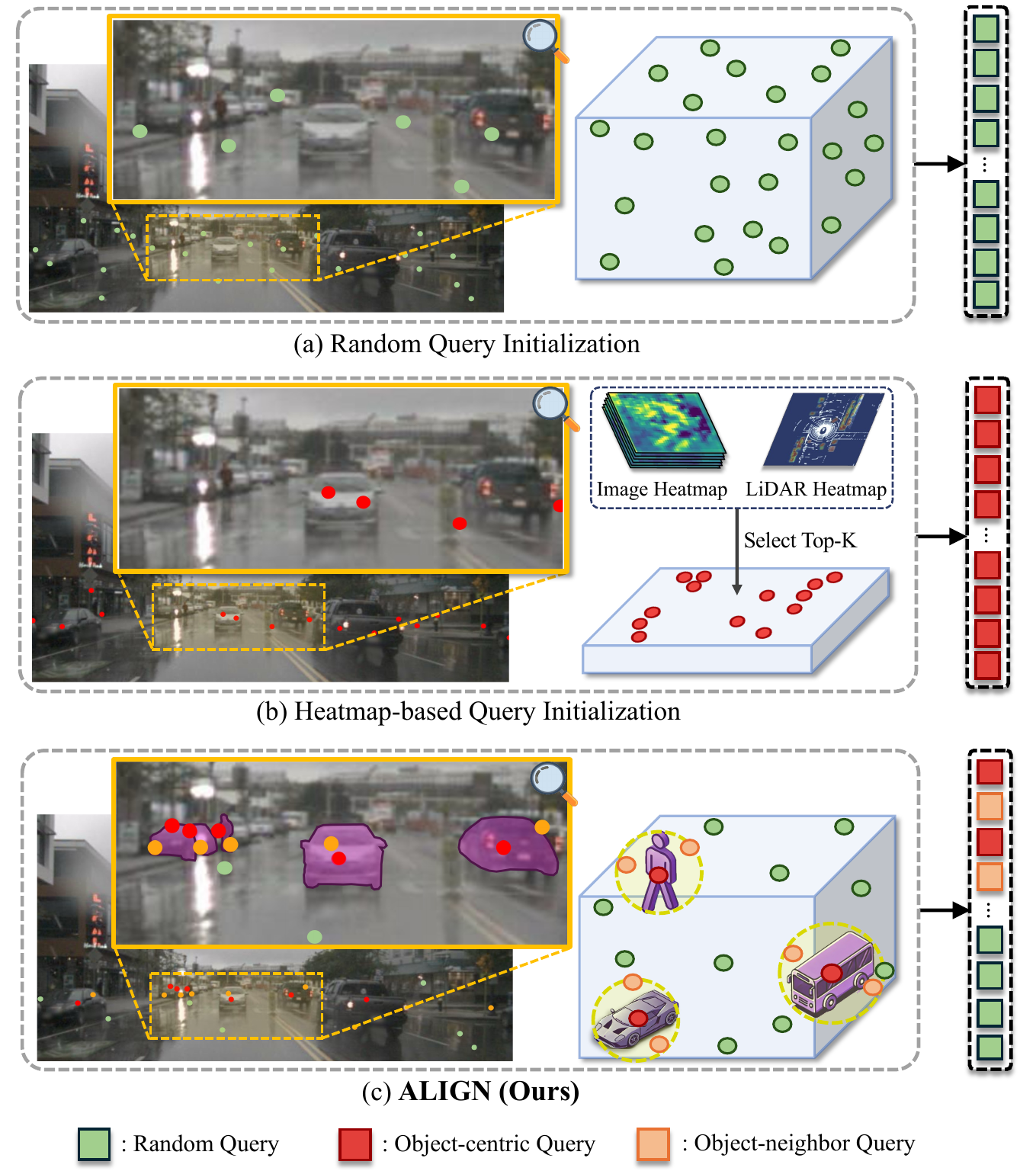}
    \caption{\textbf{Query Initialization Comparison.} Existing 3D object detectors typically adopt following query initialization schemes: (a) random initialization, where queries are sampled from uniformly distributed spatial regions, or (b) heatmap-based sampling from salient regions identified in BEV heatmaps. 
    In contrast, (c) ALIGN accurately estimates object centers and samples queries in the vicinity of each center, while maintaining balanced background queries. 
    }
    \label{fig:teaser}
    \vspace{-2em}
\end{figure}

Multi-sensor fusion based 3D object detection is a fundamental capability in autonomous driving and robotics, as it directly influences an agent’s ability to perceive its environment, identify obstacles, and navigate safely in complex urban environments.
Combining multiple sensor modalities, particularly LiDAR and cameras, has emerged as a preferred strategy to enhance detection robustness and accuracy~\cite{liang_deep_2018,liu_bevfusion_2022_arxiv,bai_transfusion_2022,huang2024detecting,zhang2024sparselif,gong2025roadside}.
While LiDAR provides accurate geometric, cameras offer rich semantic and texture cues, improving robustness under adverse lighting, heavy clutter, and occlusions.


Recently, query-based methods~\cite{gao2022adamixer,wang2023objectquerylifting2d,zhao2024detrs} have shown strong performance by training each query to attend to potential objects in 3D space. However, their effectiveness largely depends on how queries are initialized. As shown in Fig.~\ref{fig:teaser} (a) and (b), existing approaches mainly adopt two strategies:
(1) Random sampling~\cite{he2022destr,yan_cross_2023}, which uniformly distributes queries without object awareness, leading to inefficient utilization, limited coverage, and slow convergence.
(2) BEV heatmap-based sampling~\cite{liu_bevfusion_2022_arxiv,xie2023sparsefusion}, which allocates queries to top-K salient regions inferred from camera or LiDAR BEV heatmaps.

Since heatmaps primarily reflect strong visible responses, small or occluded objects often receive low confidence and are excluded.
Consequently, both strategies struggle in complex urban scenes with large variation in scale and visibility.

To address these limitations, we introduce ALIGN, a novel object-aware query initialization strategy that aligns 2D image semantics with 3D LiDAR geometry to infer spatially and semantically consistent object centers (see Fig.~\ref{fig:teaser}(c)).
Unlike BEV heatmap-based methods that infer query locations from coarse top-view confidence maps, ALIGN combines segmentation-guided projections with LiDAR-based clustering, enabling precise and robust query placement even for occluded, distant, or small objects.

ALIGN consists of three complementary modules for query initialization:
(1) Occlusion-aware Center Estimation (OCE) estimates object centers by projecting LiDAR points onto image segmentation maps, combining geometric and semantic cues for localization under occlusion.
(2) Adaptive Neighbor Sampling (ANS) enhances spatial coverage via LiDAR clustering and neighbor sampling within segmentation-consistent regions, enabling robust detection in crowded scenes.
(3) Dynamic Query Balancing (DQB) preserves global coverage by retaining a limited number of background queries, preventing overfitting to dense object regions.
These modules form a unified framework that improves localization accuracy and spatial coverage for more robust 3D object detection across diverse driving scenarios. 

We evaluate our method on the widely used nuScenes benchmark~\cite{caesar_nuscenes_2020}, a large-scale autonomous driving dataset with multi-view cameras and LiDAR sensors.
Since our approach integrates seamlessly into existing query-based detectors, we validate its generality across several state-of-the-art models, including CMT~\cite{yan_cross_2023}, TransFusion~\cite{bai_transfusion_2022}, SparseFusion~\cite{xie2023sparsefusion}, and FUTR3D~\cite{chen_futr3d_2022}.
Our model-agnostic design enables easy integration into a wide range of query-based detectors without modifying their core architectures.
Extensive experiments show consistent improvements in both mAP and NDS, with particularly strong gains for occluded and small objects in crowded urban scenes.

Our main contributions are summarized as follows:

\begin{itemize}
    \item We present ALIGN, a novel object-aware query initialization framework for multi-sensor 3D object detection that explicitly aligns 2D image semantics with 3D LiDAR geometry, enabling structured and object-centric query generation.
    \item We introduce three complementary modules (occlusion-aware center estimation, adaptive neighbor sampling, and dynamic query balancing) that collectively improve query localization, spatial coverage, and distribution balance across diverse driving scenarios.
    \item We demonstrate that ALIGN can be seamlessly integrated into various state-of-the-art query-based 3D object detectors, consistently yielding significant performance gains on the nuScenes benchmark, particularly under occlusion and crowding conditions.
\end{itemize}

\section{Related Work}
\label{sec:related}
\begin{figure*}[t]
    \centering
    \includegraphics[width=1\textwidth]{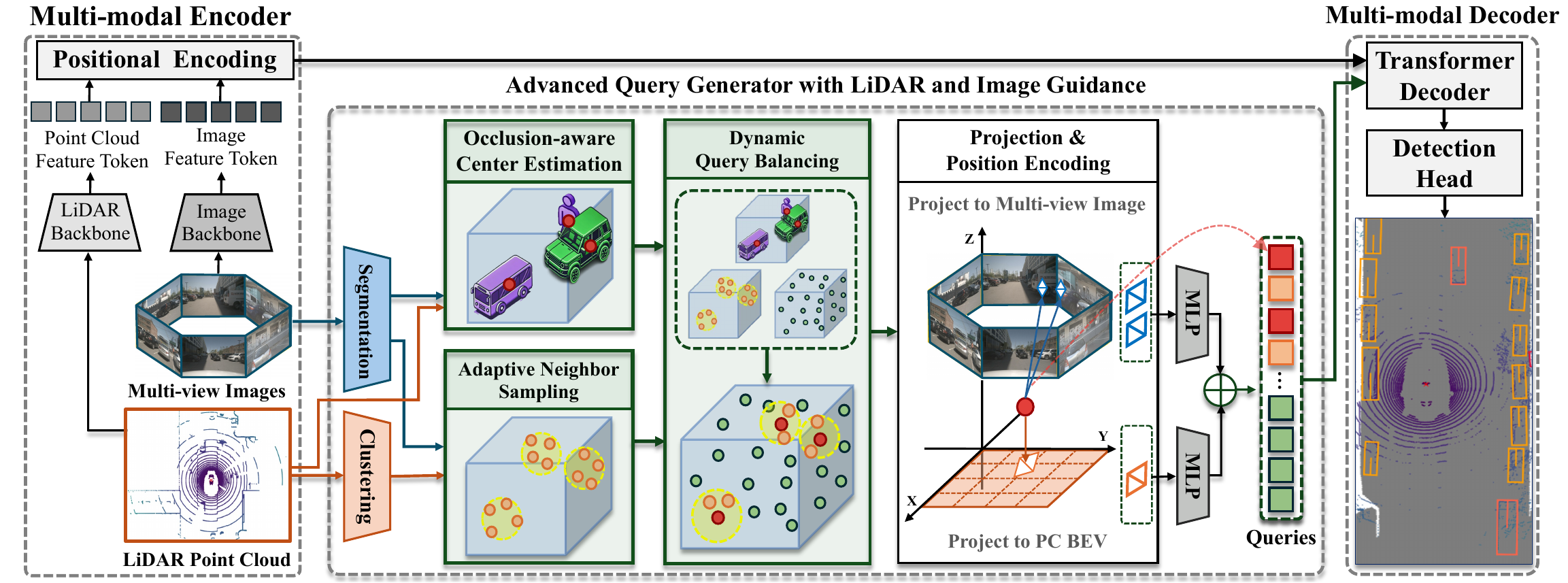}
    \caption{
    \textbf{An overview of ALIGN}, a novel query initialization strategy for query-based 3D object detection. Our framework consists of three components: (i) Occlusion-aware Center Estimation (OCE) for object center estimation from LiDAR point cloud and image segmentation, 
    (ii) Adaptive Neighbor Sampling (ANS) for generating object candidates via LiDAR clustering and sampling object-aware queries, and (iii) Dynamic Query Balancing (DQB) for augmenting background queries to ensure a balanced foreground–background distribution.
}
\vspace{-1em}
    \label{fig:model_overview}
\end{figure*}



Recently, query-based 3D object detectors have shown strong performance using learnable queries that interact with multi-modal features.
Their effectiveness mainly depends on how queries are initialized, which affects the model’s focus and convergence.
Most approaches adopt one of two strategies: (i) random initialization and (ii) heatmap-based initialization.

\subsection{Random Query Initialization}
Most query-based 3D object detectors adopt random initialization following the DETR~\cite{carion2020end} paradigm.
DETR3D~\cite{wang2022detr3d} samples 3D reference points and aggregates image features via projection.
PETR~\cite{liu_petr_2022} introduces 3D position encoding into image features for implicit spatial grounding.
FUTR3D~\cite{chen_futr3d_2023} fuses image and LiDAR features at each query via a unified transformer without BEV or anchor priors.
CMT~\cite{yan_cross_2023} improves modality alignment via coordinate-aware position embeddings.

While simple and generalizable, random initialization leads to suboptimal query placement and slow convergence due to the lack of spatial and semantic priors.
As a result, many queries fall into background regions, producing weak object cues and redundant attention.
During early training, only a few queries receive positive supervision via Hungarian matching, while others remain unaligned or collapse onto the same salient regions.
This inefficient allocation slows convergence and limits the discovery of small or occluded objects.

\subsection{Heatmap-based Query Initialization.}

To mitigate this issue, recent approaches adopt BEV heatmap-based initialization by sampling top-$k$ salient regions.
CenterFormer~\cite{zhou_centerformer_2022} selects object centers from BEV heatmaps and uses their voxel features as queries.
TransFusion~\cite{bai_transfusion_2022} generates image-guided BEV heatmaps to initialize queries.
SparseFusion~\cite{xie2023sparsefusion} samples queries from both image and LiDAR heatmaps, jointly guided by geometry and semantics.
EfficientQ3M~\cite{van2024multimodal} combines BEV heatmaps with visual cues for query placement.

Although these methods improve query efficiency by focusing on likely object regions, they rely on strong visual or geometric responses in the BEV space.
As a result, small or occluded objects with weak activations are often underrepresented, leading to missed detections in crowded scenes.

To address these limitations, we propose an object-aware and scene-adaptive query initialization strategy that leverages 3D LiDAR geometry and 2D image segmentation for robust 3D object detection.
Our method estimates object centers via segmentation-guided projection and LiDAR clustering, and samples neighboring anchors within semantically consistent regions to capture local context.
A fixed proportion of background anchors is retained to preserve global coverage and prevent overfitting to dense object areas.
This design enables precise localization while maintaining scene coverage, improving detection of small and occluded objects.
\section{Method}
\label{sec:method}

The overall framework of ALIGN is illustrated in Fig.~\ref{fig:model_overview}.  
Given multi-view images and a LiDAR point cloud, ALIGN replaces heuristic query initialization with a structured generator that aligns geometric and semantic cues prior to transformer decoding.
Specifically, ALIGN produces three types of query anchors: (1) Object-centric anchors from OCE, which estimate 3D centers by projecting segmentation-aligned LiDAR points with class-specific depth offsets. (2) Neighbor anchors from ANS, which sample semantically consistent points around clustered regions to recover occluded or truncated objects and (3) background anchors from DQB, which dynamically balance query allocation based on scene density.
All anchors are encoded in image and BEV spaces and passed to the multi-modal transformer decoder, enabling improved spatial coverage and object localization.


\begin{figure*}[t]
    \centering
    \includegraphics[width=0.96\textwidth]{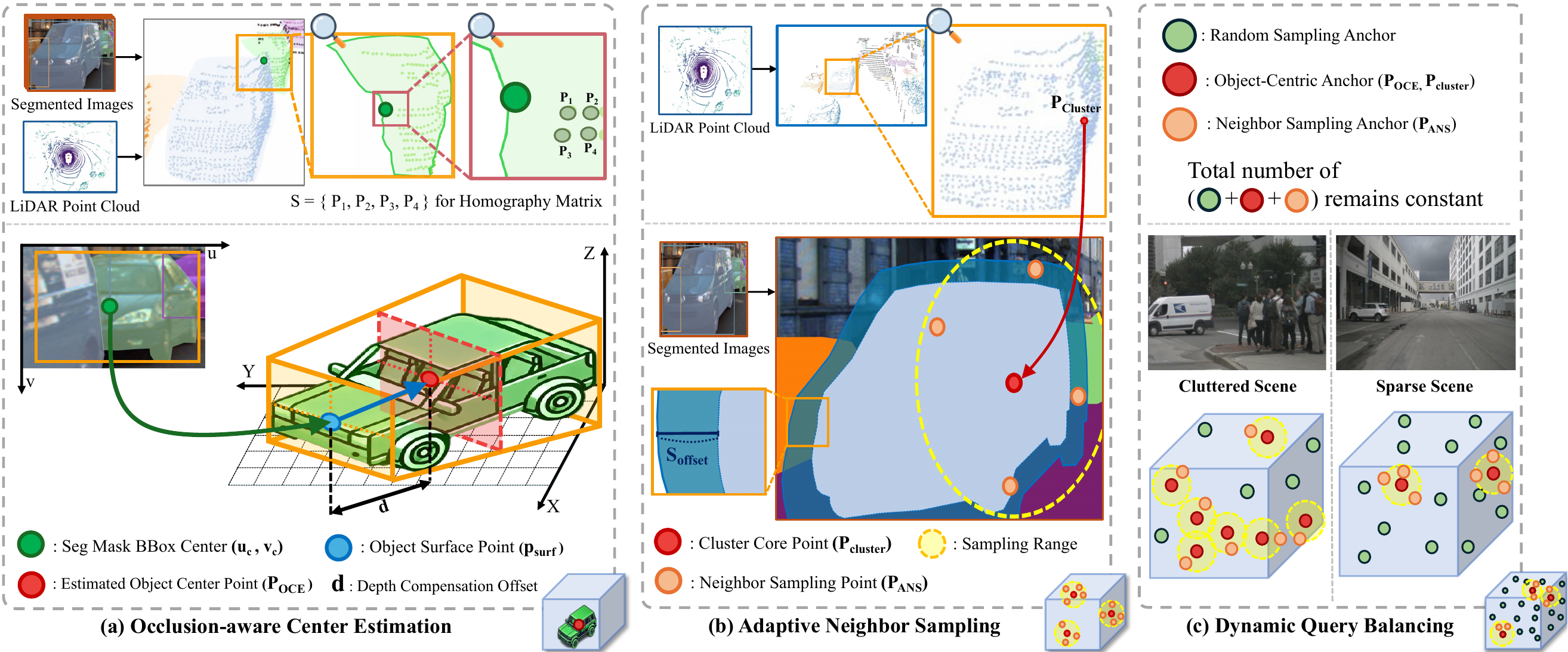}

    \caption{\textbf{Detailed illustration of each modules.}
    (a) OCE estimates object centers using LiDAR geometry and image semantics for robust localization under occlusion.
    (b) ANS samples segmentation-aligned neighbors around each LiDAR cluster core to expand anchor coverage.
    (c) DQB balances object-centric and background queries by scene complexity to preserve spatial coverage.
   }
   \vspace{-1.0em}
    \label{fig:method}
\end{figure*}

\subsection{Occlusion-Aware Center Estimation}
\label{sec:OCE}

To address the limitations of existing query initialization methods in localizing occluded, small, or sparsely sampled objects, we propose an Occlusion-Aware Center Estimation (OCE) module (Fig.~\ref{fig:method}(a)).
OCE estimates 3D object centers by combining LiDAR geometry with image semantics at the point level.
LiDAR points are projected onto image segmentation maps to associate each point with its visible instance, and the 2D–3D geometric relationship is used to estimate an approximate surface position.
A class-specific depth offset is then applied to correct surface bias and obtain a more accurate 3D center, which serves as a stable, occlusion-aware anchor for query initialization.
More implementation details and geometric interpretations of OCE are provided in the supplementary material.

3D LiDAR points \((x,y,z)\) are projected onto the multi-view image planes, yielding image coordinates \((u,v)\) via known calibration parameters. 
Let \(\mathcal{M}_{i,j}\) denote the segmentation mask of the \(j\)-th object in the \(i\)-th view. 
We select only the LiDAR points whose projections fall inside the segmentation mask \(\mathcal{M}_{i,j}\) and define the set as \(\mathbf{P}_{i,j}\) :
\begin{equation}
\mathbf{P}_{i,j} = \left\{ (u, v, x, y, z) \;\big|\, (u, v) \in \mathcal{M}_{i,j} \right\}
\label{eq:save_coords}
\end{equation}
We select four LiDAR points closest to the center \(\mathbf(u_c,v_c)\) of the 2D bounding box in image space to construct the set \(\mathbf{S}_{i,j}\) :
\begin{equation}
\mathcal{S}_{i,j} = \left\{ \operatorname{Top}_4 \left( \mathbf{P}_{i,j},\; \| (u,v) - (u_c,v_c) \|_2 \right) \right\}
\end{equation}
The bounding box center \(\mathbf(u_c,v_c)\) is provided together with the segmentation mask \(\mathcal{M}_{i,j}\) from the segmentation model.
Using these, we estimate a local homography matrix \(\mathbf{H}_{i,j}\) that maps image coordinates to LiDAR space:
\begin{equation}
[x, y, z]^\top = \mathbf{H}_{i,j} [u, v, 1]^\top.
\end{equation}
Applying \(\mathbf{H}_{i,j}\) to the 2D center \((u_c, v_c)\) yields an approximate 3D points on the object surface, denoted as \(\mathbf{p}_{\text{surf}}\) :
\begin{equation}
\mathbf{p}_{\text{surf}} = \mathbf{H}_{i,j} [u_c, v_c, 1]^\top.
\label{eq:surf_center}
\end{equation}
Although \(\mathbf{p}_{\text{surf}}\) corresponds to the visible surface, it does not coincide with the true geometric center due to its inherent surface bias.
To compensate for this, we apply a class-specific \textit{depth compensation offset} \(\mathbf{d}\), which approximates the displacement between the visible surface and the geometric center along the depth axis.
This offset is applied along the LiDAR ray, yielding the final estimated center set:
\begin{equation}
\mathbf{P}_{\text{OCE}} = \left\{\,\mathbf{p}_{\text{surf}} + d \cdot \frac{\mathbf{p}_{\text{surf}}}{\|\mathbf{p}_{\text{surf}}\|_2}\,\right\}.
\end{equation}

\subsection{Adaptive Neighbor Sampling}
\label{sec:ANS}

Although OCE reliably estimates object centers under occlusion, it depends on visibility and segmentation quality.
When objects are heavily occluded or adjacent instances have unclear boundaries, segmentation masks may be missing or merged, leading to ambiguous localization.
To address this, we introduce Adaptive Neighbor Sampling (ANS) (Fig.~\ref{fig:method}(b)), which leverages LiDAR clustering to identify object regions and samples neighboring points around each cluster core as query anchors.
We apply DBSCAN~\cite{schubert2017dbscan} to cluster LiDAR points and use each cluster core \(\mathbf{p}_{\text{cluster}}\) as an anchor.
Following Deformable DETR~\cite{zhu_deformable_2021}, we randomly sample \(N\) neighbor points within a range \(r\) around each cluster core to capture object spatial extent.
\begin{equation}
\mathbf{P}_{\text{nbr}} = \Bigl\{\, \mathbf{p}_i \;\big|\; \operatorname{dist}(\mathbf{p}_i,\mathbf{p}_{\text{cluster}}) < r,\; i=1,\dots,N \Bigr\}.
\end{equation}
This naïve sampling include irrelevant regions such as walls or the ground.
To mitigate this, we retain only points whose projections fall within a offset \(\mathbf{S_{\text{offset}}}\) of the segmentation mask \(\mathcal{M}_i\), forming the refined sampling set \(\mathbf{P}_{\text{ANS}}\):

\begin{equation}
\mathbf{P}_{\text{ANS}} = \Bigl\{\, \mathbf{p}_i \in \mathbf{P}_{\text{nbr}} \;\big|\; \operatorname{dist}\bigl(\Pi(\mathbf{p}_i), \mathcal{M}_{i}\bigr) < S_{\text{offset}} \Bigr\}.
\label{eq:padapt_def}
\end{equation}
If the number of retained points in \(\mathbf{P}_{\text{ANS}}\) is insufficient, the sampling steps are repeated up to three times to ensure sufficient coverage.
ANS complements OCE through geometry-based clustering and semantic filtering, ensuring reliable query initialization even when segmentation is incomplete or ambiguous, and producing the core anchor set \(\mathbf{p}_{\text{cluster}}\) and the refined neighbor set \(\mathbf{P}_{\text{ANS}}\).
Detailed implementation of the DBSCAN-based LiDAR clustering and parameter settings used in ANS is provided in the supplementary material.

\subsection{Dynamic Query Balancing}
\label{sec:DQB}

While OCE and ANS effectively generate object-centric anchors, assigning all queries to these regions may cause overfitting and limit generalization when some objects are missed.
On the other hand, excessive background queries lead to redundant sampling in sparse areas with limited supervision.
To resolve this trade-off, we introduce Dynamic Query Balancing (DQB) (Fig.~\ref{fig:method}(c)), which adaptively allocates remaining queries between object-centric and background anchors based on scene complexity.
After assigning object-centric anchors, the remaining queries are computed as :

\begin{equation}
N_{\text{ANS}}   = \mathcal{r}_{\text{bal}}\bigl(N_{\text{total}}-N_{\text{OCE}}-N_{\text{cluster}}\bigr), \qquad
\end{equation}
\begin{equation}
N_{\text{rand}} = (1-\mathcal{r}_{\text{bal}})\bigl(N_{\text{total}}-N_{\text{OCE}}-N_{\text{cluster}}\bigr),
\end{equation}
where \(N_{\text{total}}\) denotes the total query budget, \(N_{\text{OCE}}\) and \(N_{\text{cluster}}\) denote the numbers of anchors from OCE and ANS cluster cores.
\(\mathcal{r}_{\text{bal}}\!\in[0,1]\) is a fixed hyperparameter balancing object-centric refinement and background exploration.
The remaining queries are divided into neighbor queries \(N_{\text{ANS}}\) and background queries \(N_{\text{rand}}\).
Neighbor queries are evenly allocated across object candidates, yielding fewer samples per object in crowded scenes to preserve background coverage and denser sampling in sparse scenes to improve localization.

\begin{table*}[t]
\centering
\small
\setlength{\tabcolsep}{6pt}
\renewcommand{\arraystretch}{1.03}
\setlength{\aboverulesep}{0.5pt}
\setlength{\belowrulesep}{0.5pt}

\begin{tabularx}{\textwidth}{l | ccc | cc | cc}
\toprule
\multirow{2}{*}{\textbf{Method}} &
\multicolumn{3}{c|}{\textit{Query Init.}} &
\multicolumn{2}{c|}{\textit{Validation}} &
\multicolumn{2}{c}{\textit{Test}} \\
\cmidrule(lr){2-4}\cmidrule(lr){5-6}\cmidrule(lr){7-8}
& \textbf{Random} & \textbf{Heatmap} & \textbf{ALIGN}
& \textbf{mAP$\uparrow$} & \textbf{NDS$\uparrow$}
& \textbf{mAP$\uparrow$} & \textbf{NDS$\uparrow$} \\
\midrule
UVTR \cite{li_unifying_2022}                      & \checkmark &          &          & 63.9 & 69.7 & 65.4 & 70.4 \\
Transfusion \cite{bai_transfusion_2022}           &           & \checkmark &          & 67.3 & 70.9 & 68.9 & 71.6 \\
CMT-small \cite{yan2023crossmodaltransformerfast} & \checkmark &          &          & 67.4 & 70.2 & 68.8 & 71.6 \\
FUTR3D \cite{chen_futr3d_2023}                    & \checkmark &          &          & 67.4 & 70.9 & 69.4 & 72.1 \\
EfficientQ3M \cite{van2024multimodal}             &           & \checkmark &          & 70.5 & 72.6 & 71.2 & 73.5 \\
SparseFusion \cite{xie2023sparsefusion}           &           & \checkmark &          & 70.5 & 72.8 & 72.0 & 73.8 \\
FocalFormer3D \cite{chen2023focalformer3d}        &           & \checkmark &          & 70.5 & 73.1 & 71.6 & 73.9 \\
CMT-Large \cite{yan2023crossmodaltransformerfast} & \checkmark &          &          & 70.3 & 72.9 & 72.0 & 74.1 \\
\midrule
Transfusion + ALIGN                               &          &          & \checkmark & 67.7 {\scriptsize (+0.4)} & 71.3 {\scriptsize (+0.4)} & -- & -- \\
CMT-small + ALIGN                                 &          &          & \checkmark & 68.3 {\scriptsize (+0.9)} & 71.4 {\scriptsize (+1.2)} & 69.3 {\scriptsize (+0.5)} & 72.2 {\scriptsize (+0.6)} \\
FUTR3D + ALIGN                                    &          &          & \checkmark & 68.2 {\scriptsize (+0.8)} & 71.8 {\scriptsize (+0.9)} & 69.9 {\scriptsize (+0.5)} & 72.6 {\scriptsize (+0.5)} \\
SparseFusion + ALIGN                              &          &          & \checkmark & 70.9 {\scriptsize (+0.4)} & 73.1 {\scriptsize (+0.3)} & -- & -- \\
CMT-Large + ALIGN                                 &          &          & \checkmark &
\textbf{71.1} {\scriptsize (+0.8)} &
\textbf{73.3} {\scriptsize (+0.4)} &
\textbf{72.4} {\scriptsize (+0.4)} &
\textbf{74.5} {\scriptsize (+0.4)} \\
\bottomrule
\end{tabularx}

\vspace{-0.5em}
\caption{Performance comparison on the nuScenes dataset.
Replacing the original initialization with ALIGN consistently improves mAP and NDS across all baselines.
The best results are highlighted in \textbf{bold}. All results are reported without test-time augmentation.}
\vspace{-0.5em}
\label{tab:main_result}
\end{table*}

\subsection{Loss Functions.}
\label{sec:loss}
We employ a simple loss composed of classification and 3D bounding box regression.
Set prediction is optimized via Hungarian matching~\cite{kuhn_hungarian_1955}, with Focal loss~\cite{lin_focal_2017} \(L_{\text{cls}}\) for classification and L1 loss \(L_{\text{reg}}\) for box regression.
The final loss is the weighted sum, as shown in Eq.~\ref{eq:loss}.
In all experiments, we set\(\lambda_1=2.0\) and \(\lambda_2=0.25\).
\begin{equation}
L_{\text{total}} = \lambda_1\,L_{\text{cls}}(\mathbf{c},\hat{\mathbf{c}}) + \lambda_2\,L_{\text{reg}}(\mathbf{b},\hat{\mathbf{b}}),
\label{eq:loss}
\end{equation}
\section{Experiments}
\label{sec:experiments}

\begin{table*}[t]
\centering
\small
\setlength{\tabcolsep}{5.9pt}   
\renewcommand{\arraystretch}{1.03}
\setlength{\aboverulesep}{0.5pt}
\setlength{\belowrulesep}{0.5pt}

\begin{tabular}{l | >{\centering\arraybackslash}p{1.05cm} >{\centering\arraybackslash}p{1.35cm} | cccccccccc | c}
\toprule
\multirow{2}{*}{\textbf{Method}} &
\multicolumn{2}{c|}{\textbf{Visibility}} &
\multicolumn{10}{c|}{\textbf{mAP (\%) $\uparrow$}} &
\multirow{2}{*}{\textbf{Overall}} \\
\cmidrule(lr){2-3}\cmidrule(lr){4-13}
& Level & Num Obj. &
Car & Truck & Bus & Tra. & Const. & Ped. & Moto. & Bike & Cone & Bar. & \\
\midrule

CMT & \multirow{2}{*}{\makecell[c]{1\\{\scriptsize(0--40\%)}}} &
\multirow{2}{*}{\makecell[c]{23,437\\{\scriptsize(19.2\%)}}} &
49.7 & 8.40 & 15.5 & 17.6 & 4.30 & 56.8 & 41.7 & 19.0 & 20.5 & 25.6 & 25.9 \\
CMT + ALIGN & & &
\textbf{51.1} & \textbf{9.97} & \textbf{18.3} & \textbf{17.8} & \textbf{4.75} &
\textbf{57.6} & \textbf{44.1} & \textbf{21.3} & \textbf{22.4} & \textbf{25.8} &
\textbf{27.3} {\scriptsize (+1.4)} \\
\midrule

CMT & \multirow{2}{*}{\makecell[c]{2\\{\scriptsize(40--60\%)}}} &
\multirow{2}{*}{\makecell[c]{12,779\\{\scriptsize(10.5\%)}}} &
64.8 & 15.8 & 43.3 & 17.6 & 12.3 & 64.0 & 48.1 & 29.4 & 15.7 & 19.6 & 33.1 \\
CMT + ALIGN & & &
\textbf{66.0} & \textbf{16.9} & \textbf{45.7} & \textbf{17.9} & \textbf{12.5} &
\textbf{64.9} & \textbf{50.1} & \textbf{30.2} & \textbf{16.0} & \textbf{21.0} &
\textbf{34.1} {\scriptsize (+1.0)} \\
\midrule

CMT & \multirow{2}{*}{\makecell[c]{3\\{\scriptsize(60--80\%)}}} &
\multirow{2}{*}{\makecell[c]{17,815\\{\scriptsize(14.6\%)}}} &
79.3 & 40.4 & 65.5 & 31.2 & 25.2 & 71.0 & 73.5 & 48.5 & 29.2 & 44.2 & 50.8 \\
CMT + ALIGN & & &
\textbf{79.9} & \textbf{40.9} & \textbf{66.3} & \textbf{32.0} & \textbf{25.3} &
\textbf{71.7} & \textbf{74.7} & \textbf{50.1} & \textbf{30.3} & \textbf{45.5} &
\textbf{51.8} {\scriptsize (+1.0)} \\
\midrule

CMT & \multirow{2}{*}{\makecell[c]{4\\{\scriptsize(80--100\%)}}} &
\multirow{2}{*}{\makecell[c]{67,830\\{\scriptsize(55.7\%)}}} &
93.5 & 67.1 & 81.4 & 51.4 & 41.4 & 92.7 & 87.5 & 83.6 & 82.8 & 74.9 & 75.6 \\
CMT + ALIGN & & &
\textbf{93.8} & \textbf{67.2} & \textbf{81.6} & \textbf{52.9} & \textbf{42.5} &
\textbf{93.1} & \textbf{88.0} & \textbf{84.9} & \textbf{83.7} & \textbf{75.5} &
\textbf{76.3} {\scriptsize (+0.7)} \\
\midrule

CMT & \multirow{2}{*}{\makecell[c]{Total\\{\scriptsize(0--100\%)}}} &
\multirow{2}{*}{\makecell[c]{121,861\\{\scriptsize(100\%)}}} &
88.7 & 65.4 & 78.3 & 46.6 & 33.9 & 87.7 & 79.7 & 70.1 & 79.5 & 72.9 & 70.3 \\
CMT + ALIGN & & &
\textbf{89.2} & \textbf{65.8} & \textbf{79.0} & \textbf{47.0} & \textbf{34.8} &
\textbf{88.2} & \textbf{80.9} & \textbf{71.7} & \textbf{81.0} & \textbf{73.5} &
\textbf{71.1} {\scriptsize (+0.8)} \\
\bottomrule
\end{tabular}

\vspace{-0.5em}
\caption{Detection performance across visibility levels on the nuScenes validation set using CMT-Large.}
\label{tab:occlusion_level_comparison}
\vspace{-1.0em}
\end{table*}

\subsection{Implementation and Evaluation Details.}
\paragraph{Dataset and Metrics.}
We use nuScenes dataset~\cite{caesar_nuscenes_2020} to evaluate our method’s effectiveness.
The nuScenes dataset includes six cameras, five radars, and one LiDAR, offering a 360-degree field of view.
The dataset contains 1{,}000 scenes split into 700/150/150 for training/validation/test and includes $\sim$1.4M annotated 3D bounding boxes covering 10 common object classes found in driving environments.
We report mean Average Precision (mAP) and nuScenes Detection Score (NDS) as the main metrics.

\paragraph{Implementation Details.}
We implement ALIGN on five query-based LiDAR-camera detectors and retain each baseline’s original backbone and training settings for fair comparison.
Segmentation masks are generated by Mask R-CNN pretrained on nuImages, sharing the image backbone and performing FP16 inference at $800{\times}320$ resolution.
The region of interest is set to $[-54,54]$ m in $X/Y$ and $[-5,3]$ m in $Z$, and all feature and query embeddings use dimension $D=256$.
All models are trained for 20 epochs on 4$\times$A6000 Ada GPUs using AdamW~\cite{loshchilov2019decoupledweightdecayregularization} with a cyclic learning rate policy~\cite{smith_cyclical_2017} and weight decay of $1{\times}10^{-2}$.
We employ CBGS~\cite{zhu_class-balanced_2019} and AutoAlign~\cite{chen2022deformable} to address class imbalance.

\subsection{Experimental Results}
\paragraph{Quantitative Results.}
Table~\ref{tab:main_result} reports the performance of five query-based LiDAR-camera detectors on the nuScenes validation and test sets before and after applying ALIGN.
Across all baselines, ALIGN consistently improves both mAP and NDS, demonstrating its general applicability.
The largest gains are observed when ALIGN replaces random initialization, which lacks spatial priors, with semantically grounded and geometry-aware object anchors.
For example, CMT-small improves from 67.4 to 68.3 mAP (+0.9) and from 70.2 to 71.4 NDS (+1.2), while FUTR3D achieves +0.8 mAP and +0.9 NDS.
In contrast, heatmap-guided methods such as TransFusion and SparseFusion show smaller yet consistent gains (+0.4 mAP / +0.4 NDS), as they already leverage BEV-based spatial cues during initialization.
The best-performing model, CMT-Large, reaches 71.1 mAP and 73.3 NDS, confirming that ALIGN scales effectively to larger architectures.
Overall, replacing heuristic or random initialization with structured, semantically grounded anchors leads to more reliable query allocation and consistent performance gains.
Additional segmentation robustness analysis is provided in the supplementary material.

\paragraph{Performance across Visibility Levels.}
The nuScenes benchmark categorizes objects into four visibility levels, from Level~1 (0--40\%, heavily occluded) to Level~4 (80--100\%, fully visible), based on 2D bounding box visibility.
In the validation set, 19.2\% of all instances fall into Level~1, indicating that heavily occluded objects are common in urban scenes.
Table~\ref{tab:occlusion_level_comparison} reports class-wise mAP across visibility levels using CMT-Large as the baseline to analyze the impact of ALIGN under varying visibility conditions.
As expected, baseline performance degrades as visibility decreases (Level~4 $\rightarrow$ Level~1), while replacing the original query initialization with ALIGN consistently improves mAP at all levels, with larger gains under heavier occlusion (e.g., +1.4 at Level~1 vs.\ +0.7 at Level~4).
Notably, significant improvements are observed for small and thin categories such as \textit{Bicycle} (+1.6) and \textit{Traffic Cone} (+2.2) at low visibility, highlighting the practical benefit of ALIGN in urban scenes.

\begin{table}[t]
\centering
\small
\setlength{\tabcolsep}{3.9pt}
\renewcommand{\arraystretch}{1.15}
\setlength{\aboverulesep}{0.6pt}
\setlength{\belowrulesep}{0.6pt}

\begin{tabular}{c | c c c | c c | c c}
\toprule
 & \textbf{OCE} & \textbf{ANS} & \textbf{DQB} 
 & \textbf{mAP$\uparrow$} & \textbf{NDS$\uparrow$}
 & \textbf{Lat.{\scriptsize (ms)}} & \textbf{Mem.{\scriptsize (MB)}} \\
\midrule
(a) &  &  &  & 67.4 & 70.2 & 163 & 2812 \\

\midrule
(b) & \checkmark &  &  & 68.0 & 70.8 & 190 & 4066 \\
(c) &  & \checkmark &  & 67.6 & 70.5 & 187 & 3084 \\
(d) & \checkmark & \checkmark &  & 68.2 & 71.2 & 215 & 4360 \\

\midrule
(e) & \checkmark & \checkmark & \checkmark 
    & \textbf{68.3} 
    & \textbf{71.4} 
    & 215 
    & 4360 \\
\bottomrule
\end{tabular}

\vspace{-0.5em}
\caption{Ablation of ALIGN modules on CMT-small}
\label{tab:ablation}
\vspace{-0.6em}
\end{table}

\begin{table}[t]
\vspace{-0.1em}
\centering
\small
\renewcommand{\arraystretch}{1.0}
\setlength{\tabcolsep}{4pt}
\begin{tabular*}{\linewidth}{@{\extracolsep{\fill}} c | cc}
\toprule
Depth Compensation Offset \textit{d} & \textbf{mAP $\uparrow$} & \textbf{NDS $\uparrow$} \\
\midrule
-- & 67.4 & 70.2 \\
{[0,0,0,0,0,0,0,0,0,0]} & 67.9 & 70.5 \\
{[1,2,2,2,2,0,0,0,0,0]} & \textbf{67.9} & 70.6 \\
{[1.5,3,3,3,3,0,0,0,0,0]} & \textbf{68.0} & \textbf{70.8} \\
\bottomrule
\end{tabular*}
\vspace{-0.5 em}
\caption{Ablation of depth offsets in OCE.}
\vspace{-1.0 em}
\label{tab:ablation_OCE}
\end{table}


\begin{table}[t]
\centering
\small
\setlength{\tabcolsep}{4pt}
\renewcommand{\arraystretch}{1.1}

\begin{tabular*}{\linewidth}{@{\extracolsep{\fill}} c c | c c}
\toprule
\textbf{Sampling Range} & \textbf{Sample Offset} & \textbf{mAP $\uparrow$} & \textbf{NDS $\uparrow$} \\
\midrule
--    & --  & 67.4 & 70.2 \\
0.015 & 15  & 67.5 & 70.3 \\
0.015 & 30  & \textbf{67.6} & \textbf{70.5} \\
0.030 & 15  & 67.5 & 70.4 \\
0.030 & 30  & \textbf{67.6} & \textbf{70.5} \\
\bottomrule
\end{tabular*}

\vspace{-0.5em}
\caption{Effect of sampling range and offset in ANS.}
\label{tab:ablation_ANS}
\vspace{-0.52em}
\end{table}

\begin{table}[t]
\vspace{-0.1em}
\centering
\small
\setlength{\tabcolsep}{4pt}
\renewcommand{\arraystretch}{1.0}

\begin{tabular*}{\linewidth}{@{\extracolsep{\fill}} c c c | c c}
\toprule
$\mathcal{r}_{\text{balance}}$ &
$N_{\mathbf{ANS}}^{\text{max}}$ &
$N_{\mathbf{rand}}^{\text{max}}$ &
\textbf{mAP $\uparrow$} &
\textbf{NDS $\uparrow$} \\
\midrule
0.32 & 288 & 612 & 65.1 & 69.3 \\
0.16 & 144 & 756 & 68.1 & 71.2 \\
0.08 & 72  & 828 & \textbf{68.3} & \textbf{71.4} \\
0.04 & 54  & 846 & 68.2 & 71.2 \\
\bottomrule
\end{tabular*}

\vspace{-0.5em}
\caption{Effect of balancing parameter in DQB.}
\label{tab:ablation_DQB}
\vspace{-1.0em}
\end{table}

\paragraph{Ablation of ALIGN Modules.}
Table~\ref{tab:ablation} presents the ablation study on CMT-small using the nuScenes validation set.
Adding OCE improves performance to 68.0/70.8 (mAP/NDS) by providing more reliable, segmentation-guided center localization.
Using ANS alone reaches 67.6/70.5, indicating that adaptive sampling around LiDAR clusters partially compensates for missing center priors, though the fixed neighbor budget limits its effect.
Combining OCE and ANS further improves performance to 68.2/71.2, showing their complementary roles in precise localization and adaptive spatial coverage.
Finally, introducing DQB achieves the highest result (68.3/71.4), yielding total gains of +0.9/+1.2 over the baseline by balancing object-centric and background queries and preventing dominance by over-represented regions.
Overall, each module contributes complementary benefits, forming a unified framework that progressively improves detection performance.

\begin{figure*}[t]
    \centering
    \includegraphics[width=1.0\textwidth]{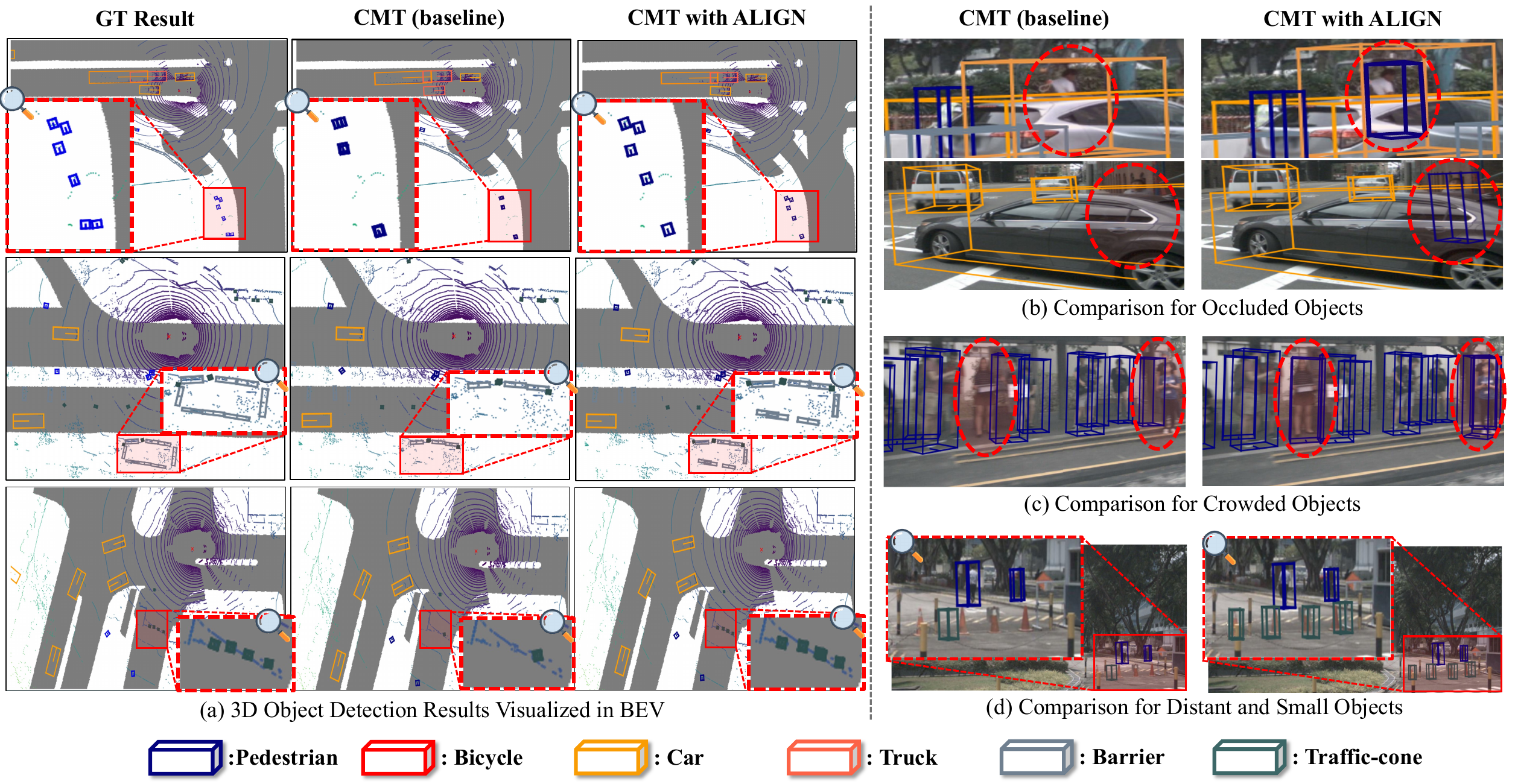}
    \caption{\textbf{Qualitative Comparison with and without ALIGN.}
    (a) BEV visualizations show that ALIGN improves localization of small and occluded objects, reducing missed detections and refining box placement.  
    (b) Our method detects heavily occluded pedestrians missed by the baseline. 
    (c) In crowded scenes, ALIGN better resolves overlaps, improving object separation and robustness. 
    (d) Detection is more accurate for distant and small objects. 
    Red dotted regions highlight areas where ALIGN outperforms the baseline.
    }
    \vspace{-1.0em}
    \label{fig:quality}
\end{figure*}




\paragraph{Depth Compensation Offset in OCE.}
In OCE, LiDAR-based center estimation relies on visible surfaces from image-guided segmentation. 
Since LiDAR observes only object surfaces, the estimated centers may deviate from the true geometric centers, especially for large or elongated objects such as trucks and construction vehicles.

To mitigate this issue, we introduce a class-specific depth compensation offset $\mathbf{d}$ along the LiDAR ray toward the object interior. 
Following QAF2D~\cite{ji2024enhancing}, the offset is derived from dataset-level 3D box size statistics of each class, providing a data-driven correction rather than heuristic tuning.
Detailed derivation and class-wise offset ranges are provided in the supplementary material.

Table~\ref{tab:ablation_OCE} reports ablation results for different offset settings.
Compared to the baseline without OCE and the zero-offset variant, applying non-zero offsets consistently improves both mAP and NDS.
Among the evaluated configurations, the dataset-informed offset $[1.5, 3, 3, 3, 3, 0, 0, 0, 0, 0]$ achieves the best performance (68.0 / 70.8 mAP / NDS) and is adopted in our model.

\paragraph{Sampling Range and Offset in ANS.}
Table~\ref{tab:ablation_ANS} reports the ablation results of the sampling range and offset in the ANS module.
The sampling range is defined as a ratio of the Region of Interest (ROI) within \([-54, 54]\) m, where 0.030 corresponds to approximately 3.24\,m and determines the spatial extent for neighbor selection around each LiDAR cluster.
A range of 0.015 already provides sufficient local coverage, and increasing it to 0.030 yields only marginal improvement, indicating that most relevant geometry is captured within a compact neighborhood.
The offset controls how many pixels beyond the segmentation mask are included during neighbor aggregation.
Larger offsets consistently improve mAP and NDS by slightly expanding the sampling boundary, allowing valid points near object boundaries to be included and leading to more reliable query initialization for partially occluded or irregularly shaped objects.

\paragraph{Balancing Parameter in DQB.}
Table~\ref{tab:ablation_DQB} analyzes the effect of the balancing coefficient $\mathcal{r}_{\text{balance}}$ in the DQB module, where a total of 900 queries is used.
After assigning OCE-estimated centers and ANS cluster cores as query anchors, DQB allocates the remaining queries between object-centric and background regions.
The coefficient $\mathcal{r}_{\text{balance}}$ controls the maximum ratio of object-centric queries by defining upper bounds $N_{\mathbf{ANS}}^{\text{max}}$ and $N_{\mathrm{rand}}^{\text{max}}$, while the actual numbers adapt dynamically to scene density.
Performance peaks at $\mathcal{r}_{\text{balance}}=0.08$ ($N_{\mathbf{ANS}}^{\text{max}}=72$), achieving 68.3 mAP / 71.4 NDS.
Larger ratios (e.g., 0.16 or 0.32) allocate excessive object-centric queries, leading to redundant representations and reduced background coverage,
while smaller ratios (e.g., 0.04) limit the model’s ability to represent object regions effectively.
These results indicate that $\mathcal{r}_{\text{balance}}=0.08$ provides the best trade-off, maintaining balanced query allocation across varying scene complexities and stabilizing detection performance.



\paragraph{Analysis of Computational Overhead.}
We evaluate the computational overhead of ALIGN on the nuScenes validation set with CMT-small as the base detector, using a single NVIDIA A6000 Ada GPU.
Table~\ref{tab:ablation} summarizes the average inference latency per sample and GPU memory consumption.
OCE introduces moderate overhead (+27\,ms, +1254\,MB) thanks to segmentation-guided center estimation using shared image backbone features.
ANS adds minor overhead (+24\,ms, +272\,MB), as clustering and neighbor sampling operate on a sparse set of LiDAR points and are executed on the GPU.
DQB incurs negligible overhead, as it only redistributes foreground and background queries without additional feature extraction.
Overall, ALIGN achieves consistent accuracy improvements with modest computational overhead, making it suitable for practical deployment.

\paragraph{Qualitative Results.}
Figure~\ref{fig:quality} presents qualitative comparisons on the nuScenes validation set, illustrating how ALIGN improves detection robustness under diverse and challenging scenarios.  
As shown in Fig.~\ref{fig:quality}(a), BEV visualizations reveal higher localization accuracy and recall, particularly for small or partially visible objects. This improvement is attributed to ALIGN’s structured query initialization, which integrates geometric priors, adaptive spatial sampling, and balanced query allocation, leading to more accurate and spatially consistent query placement across the scene.  
Figure~\ref{fig:quality}(b) shows that ALIGN detects heavily occluded objects more reliably by leveraging segmentation-guided centers, enabling consistent center estimation even from partial surfaces and improving box placement when large portions of objects are invisible.  
In crowded scenes (Fig.~\ref{fig:quality}(c)), baseline detectors often struggle to separate adjacent instances due to ambiguous query initialization and overlapping receptive fields, which frequently leads to merged or misplaced detections. ALIGN mitigates this issue by initializing queries from well-separated object centers and restricting feature aggregation within object-specific regions, thereby preserving clear boundaries and improving instance separation.  
Finally, Fig.~\ref{fig:quality}(d) demonstrates improved detection of distant and small objects, where adaptive sampling around cluster cores compensates for sparse LiDAR observations and enhances local geometric representation in low-density regions.  
Overall, these results show that ALIGN mitigates key failure modes—occlusion, crowding, and distance—through a unified, geometry-aware query initialization framework.

\section{Conclusion}
\label{sec:conclusion}

We presented \textbf{ALIGN}, a structured query-initialization framework that integrates geometric and semantic cues for multi-modal 3D object detection. 
While prior sampling strategies either distribute queries uniformly or focus on clearly visible regions, often missing small or occluded objects, ALIGN aligns LiDAR geometry with image semantics to guide balanced and object-aware query placement across foreground and background regions. 
By integrating OCE, ANS, and DQB, ALIGN enforces spatial consistency and yields more stable query distributions with stronger cross-modal alignment.
Extensive experiments demonstrate that ALIGN consistently improves performance across multiple detectors, with particularly large gains under low-visibility conditions. 
These results show that improving the initialization stage alone can deliver reliable performance gains without modifying detector architectures.
Although ALIGN introduces additional computation and partially depends on segmentation quality in OCE, these are practical rather than fundamental limitations and can be mitigated through lightweight segmentation or self-supervised geometric estimation. 
Overall, this work takes a clear step toward more structured and object-aware query initialization, highlighting the importance of geometry–semantic alignment for robust perception in complex driving scenes.

\clearpage
\appendix
\clearpage

\begin{strip}
\centering
{\LARGE\bf Supplementary Material}\\[1.5em]
{\large ALIGN: Advanced Query Initialization with LiDAR-Image Guidance for Occlusion-Robust 3D Object Detection}
\end{strip}

\vspace{2.0em}

\section{Class-Specific Depth Offset in OCE.}

In the Occlusion-aware Center Estimation (OCE) module, object centers are initialized by projecting LiDAR points onto the image plane and computing a representative surface point \( p_{\text{surf}} \) from image-guided segmentation.
However, since LiDAR sensors only capture visible surfaces, the recovered 3D point typically lies on the front-facing surface of an object rather than at its geometric center.
This leads to a systematic depth bias, where the estimated center is consistently shifted toward the camera, especially for elongated or partially occluded objects.

To address this limitation, we introduce a depth offset \( d \) along the LiDAR ray direction, which shifts the surface point deeper into the scene to better approximate the true object center.
Crucially, this offset is not treated as a free hyper-parameter.
Instead, following~\cite{ji2024enhancing}, we derive class-specific offset ranges from dataset-level object size statistics.

Specifically, we compute the empirical distributions of object width, height, and length for each category and estimate the expected distance between the visible surface and the geometric center.
As a result, large and elongated objects require larger offsets, while compact or symmetric ones need only minimal adjustment.

Table~\ref{tab:OCE_parameter} summarizes the resulting class-wise size statistics and the derived depth offset ranges.
These values are fixed during inference and introduce no additional learnable parameters.
As shown in Table~\ref{tab:occlusion_level_comparison_NDS}, the proposed class-specific depth offsets consistently reduce Average Scale Error (ASE) and Average Orientation Error (AOE), confirming that this data-driven compensation improves the geometric quality of 3D bounding box estimation.

\begin{table}[h]
\centering
\footnotesize
\setlength{\tabcolsep}{2.3pt}
\renewcommand{\arraystretch}{1.0}
\begin{tabular}{l
    >{\scriptsize}c
    >{\scriptsize}c
    >{\scriptsize}c
    >{\scriptsize}c}
\toprule
\textbf{Category} & $(w_\mathrm{min}, w_\mathrm{max})$ & $(h_\mathrm{min}, h_\mathrm{max})$ & $(l_\mathrm{min}, l_\mathrm{max})$ & $(d_\mathrm{min}, d_\mathrm{max})$ \\
\midrule
Car           & (1.4, 2.8) & (1.2, 3.1) & (3.4, 6.6)   & (0.6, 3.9) \\
Pedestrian    & (0.3, 1.0) & (1.0, 2.2) & (0.3, 1.3)   & (0.2, 1.4) \\
Bus           & (2.6, 3.5) & (2.8, 4.6) & (6.9, 13.8)  & (1.3, 7.5) \\
Truck         & (1.7, 3.5) & (1.7, 4.5) & (4.5, 14.0)  & (0.8, 7.6) \\
Trailer       & (2.2, 2.3) & (3.3, 3.9) & (1.7, 14.0)  & (0.8, 7.4) \\
Constr. veh.  & (2.1, 3.4) & (2.0, 3.0) & (3.7, 7.6)   & (1.0, 4.4) \\
Motorcycle    & (0.4, 1.5) & (1.1, 2.0) & (1.2, 2.8)   & (0.2, 1.9) \\
Bicycle       & (0.4, 0.9) & (0.9, 2.0) & (1.3, 2.0)   & (0.2, 1.5) \\
Traffic cone  & (0.2, 1.2) & (0.5, 1.4) & (1.3, 2.0)   & (0.1, 1.4) \\
Barrier       & (1.7, 3.6) & (0.8, 1.4) & (0.3, 0.8)   & (0.2, 2.0) \\
\bottomrule
\end{tabular}
\caption{Per-class bounding box size ranges and corresponding depth offset ranges used in OCE. Each prior is computed from dataset-level statistics and reflects the expected distance between the visible surface and the true object center, helping compensate for depth bias. All values are in meters.}

\label{tab:OCE_parameter}
\end{table}

\begin{table*}[t]
\centering
\small
\begin{tabular}{c c cccccccccc c}
\toprule
\textbf{Method} & \textbf{Metric} & Car & Truck & Bus & Tra. & Const. & Ped. & Moto. & Bike & Cone & Bar. & Total \\
\midrule
CMT
 & \multirow{2}{*}{\makecell{ASE}$(\downarrow)$} & 14.0 &  \textbf{17.8} & 16.8 & 21.8 &  44.8 & 27.6 & 23.1 & 25.7 & 31.5 & 27.1 & 25.0 \\
CMT + ALIGN  &  & 14.0 &  17.9 & \textbf{16.7} & \textbf{21.7} &  \textbf{44.5} & \textbf{27.5} & \textbf{22.3} & \textbf{25.1} & \textbf{30.9} & 27.1 & \textbf{24.7} \textbf{\textsubscript{{(-0.3)}}} \\
\midrule

CMT
 & \multirow{2}{*}{\makecell{AOE}$(\downarrow)$} & 4.00 &  4.20 & 4.00 & 35.9 &  83.6 & 28.5 & 22.0 & 30.2 & -- & 4.30 & 24.1 \\
CMT + ALIGN  &  & \textbf{3.80} &  \textbf{3.80} & 4.00 & \textbf{33.8} &  \textbf{79.8} & \textbf{26.9} & \textbf{20.1} & \textbf{27.2} & -- & \textbf{4.10} & \textbf{22.6} \textbf{\textsubscript{{(-1.5)}}} \\

\bottomrule
\end{tabular}
\caption{Class-specific depth offsets improve ASE and AOE on the nuScenes validation set.}
\vspace{-1.0em}
\label{tab:occlusion_level_comparison_NDS}
\end{table*}

\section{Details of Object Center Estimation in OCE}

OCE estimates each object’s 3D center by geometrically linking image-space object cues with LiDAR measurements.
From each camera view, the segmentation network identifies object regions and computes their 2D centroids, which serve as approximate object centers in image space.
Given a 2D centroid $(u_c, v_c)$, we project it into the LiDAR coordinate system using the calibrated camera–LiDAR transformation.

\begin{figure*}[t]
  \centering
  \includegraphics[width=0.8\textwidth]{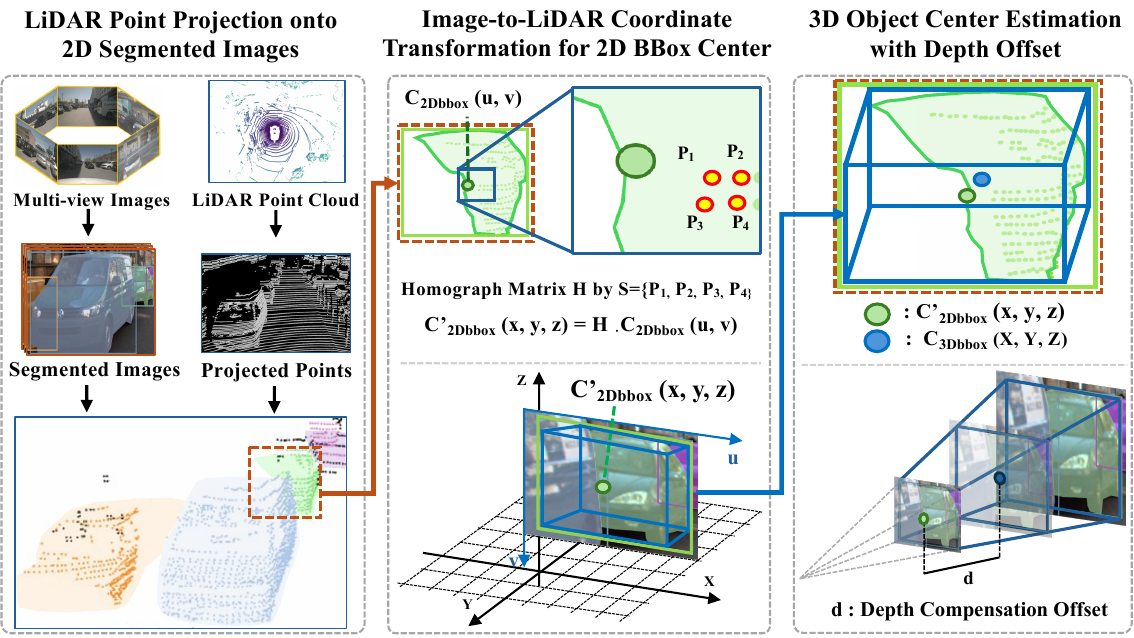}
  \vspace{-0.5em}
  \caption{
\textbf{Visualization of 3D object center estimation in OCE.}
The 2D center from segmentation is projected to the LiDAR frame and refined with a depth offset for geometry-consistent query initialization.
}
  \vspace{-0.5 em}
  \label{fig:oce_center_estimation_detail}
\end{figure*}

To obtain a 3D estimate, we associate the projected image center with a local set of LiDAR points inside the same segmented region.
Specifically, we collect a small set of neighboring LiDAR points
$\mathcal{S} = \{\mathbf{P}_1, \mathbf{P}_2, \mathbf{P}_3, \mathbf{P}_4\}$
and use them to define a local geometric mapping from the image plane to the LiDAR space.
This transformation yields a geometry-consistent estimate of the visible surface location corresponding to the image centroid, denoted as $\mathbf{C}'_{\text{2Dbbox}}$ in Fig.~\ref{fig:oce_center_estimation_detail}.

Since LiDAR observes only the visible surface of an object, the resulting 3D point typically lies on the front-facing surface rather than at the true geometric center, especially for elongated or partially occluded objects.
To compensate for this systematic depth bias, OCE applies a depth offset $d$ along the viewing direction, shifting the surface point deeper into the scene.
$\mathbf{v}$ denotes the viewing direction.
\[
\mathbf{C}_{\text{3Dbbox}} = \mathbf{C}'_{\text{2Dbbox}} + d \cdot \mathbf{v},
\]

The offset value is determined empirically from dataset-level object size statistics and remains fixed during inference.
This simple geometric correction introduces no additional learnable parameters and incurs negligible computational cost.
Despite its simplicity, it aligns the image-derived center with the LiDAR’s 3D structure, improving the spatial accuracy of query initialization and enabling robust detection of small or partially occluded objects.

\section{Details of LiDAR Clustering in ANS}

In the Adaptive Neighbor Sampling (ANS) module, object candidates are extracted by applying DBSCAN~\cite{schubert2017dbscan} to LiDAR points within a predefined region-of-interest (ROI).
We use a fixed parameters $\textit{eps}=0.6$ and a minimum cluster size $\textit{minPts}=7$ for all scenes.

These parameters follow the commonly adopted clustering settings used for nuScenes LiDAR data and are kept fixed across all experiments.
In practice, we found that ANS is not highly sensitive to moderate variations of these values, as its role is not to produce precise object segmentation but to provide coarse object proposals for query initialization.
As long as clusters roughly capture object-scale point groups, the subsequent transformer decoding can refine localization.

To accelerate clustering, we utilize the GPU-based DBSCAN implementation from NVIDIA’s cuML library~\cite{raschka2020machine}.
Since LiDAR points are already represented as PyTorch~\cite{paszke2019pytorch} tensors on the GPU, we convert them to the cuML format using DLpack~\cite{DLPack}-based zero-copy transformation.
On a single NVIDIA A6000 Ada GPU, DBSCAN clustering for one nuScenes scene takes only 27ms, achieving over $50\times$ speedup compared to the CPU version.
Although the clustering step introduces additional computation, its overhead remains small and does not affect real-time inference.

\begin{table}[t]
\centering
\renewcommand{\arraystretch}{1.0}
\setlength{\tabcolsep}{3.5pt}
\scriptsize
\begin{tabular*}{\linewidth}{@{\extracolsep{\fill}} lccccc}
\toprule
\textbf{Model} & \textbf{Lat. (ms)} & \textbf{Mem. (MB)} & \textbf{mAP} & \textbf{NDS} \\
\midrule
\textit{No segmentation (baseline)} & 163 & 2812 & 67.4 & 70.2 \\
\midrule
RDRNet-s & 237\textsubscript{ (+38)} & 4275 & 67.8\textsubscript{ (+0.4)} & 70.3\textsubscript{ (+0.1)} \\
YOLACT & 346\textsubscript{ (+147)} & 4707 & 67.7\textsubscript{ (+0.3)} & 70.2\textsubscript{ (--)} \\
Cascade Mask R-CNN & 588\textsubscript{ (+389)} & 5414 & 67.9\textsubscript{ (+0.5)} & 70.5\textsubscript{ (+0.3)} \\
SOLOv2 & 465\textsubscript{ (+266)} & 6984 & 67.9\textsubscript{ (+0.5)} & 70.5\textsubscript{ (+0.3)} \\
\midrule
Mask R-CNN & 190\textsubscript{ (+27)} & 4066 & \textbf{68.0}\textsubscript{ (+0.6)} & \textbf{70.8}\textsubscript{ (+0.6)} \\

\bottomrule
\end{tabular*}
\caption{Effect of segmentation model choice in OCE on computational cost and detection performance. 
Results are measured on the nuScenes validation set using CMT-small as the base detector.}
\vspace{-1.5 em}
\label{tab:seg_model_perf}
\end{table}

\section{Segmentation Model Comparison}

To analyze the dependence of OCE on the segmentation backbone, we evaluate several pretrained instance segmentation models from MMDetection, including Mask R-CNN~\cite{he2017mask}, Cascade Mask R-CNN~\cite{cai2019cascade}, YOLACT~\cite{bolya2019yolact}, SOLOv2~\cite{wang2020solov2}, and the real-time model RDRNet~\cite{yang2024reparameterizable}. 
All models are fine-tuned on the nuImages dataset using their released pretrained weights and original training configurations.

For a fair comparison, each model processes six multi-view images from nuScenes, and both GPU memory consumption and inference latency are measured on a single NVIDIA A6000 Ada GPU. 
Mask R-CNN shares the image backbone with the base detector as described in the main paper, while the other segmentation models are evaluated as standalone networks and fine-tuned for three epochs on nuImages for performance comparison.
As a result, their latency and memory overhead are higher, which reflects the cost of using independent segmentation backbones rather than a shared feature extractor.

Table~\ref{tab:seg_model_perf} reports the results on the nuScenes validation set using CMT-Large as the base detector.
The first row shows the baseline that applies only ANS without segmentation guidance.
Although heavier segmentation networks incur significantly higher computational cost, their detection accuracy gains remain marginal.
In contrast, the lightweight RDRNet-S achieves comparable mAP and NDS with substantially lower latency and memory usage.

Importantly, across all settings, incorporating OCE consistently improves performance over the baseline, even when lightweight or real-time segmentation models are used.
This indicates that OCE does not critically depend on highly accurate or finely detailed segmentation masks.
While precise instance segmentation is beneficial, our results show that coarse object-level masks are already sufficient to provide reliable geometric cues for center initialization.
In practice, what matters most is whether the segmentation successfully identifies the presence of an object region, rather than how precisely its boundary is delineated.

These results suggest that OCE is robust to moderate segmentation noise and remains effective as long as rough object regions can be recovered.
The main performance gains stem from geometric alignment and localized query initialization, which remain stable even under imperfect segmentation, alleviating concerns about brittleness in adverse conditions such as rain, motion blur, or distant objects.

\section{More Visualization Results}

\begin{figure*}[t]
  \centering
  \includegraphics[width=0.8\textwidth]{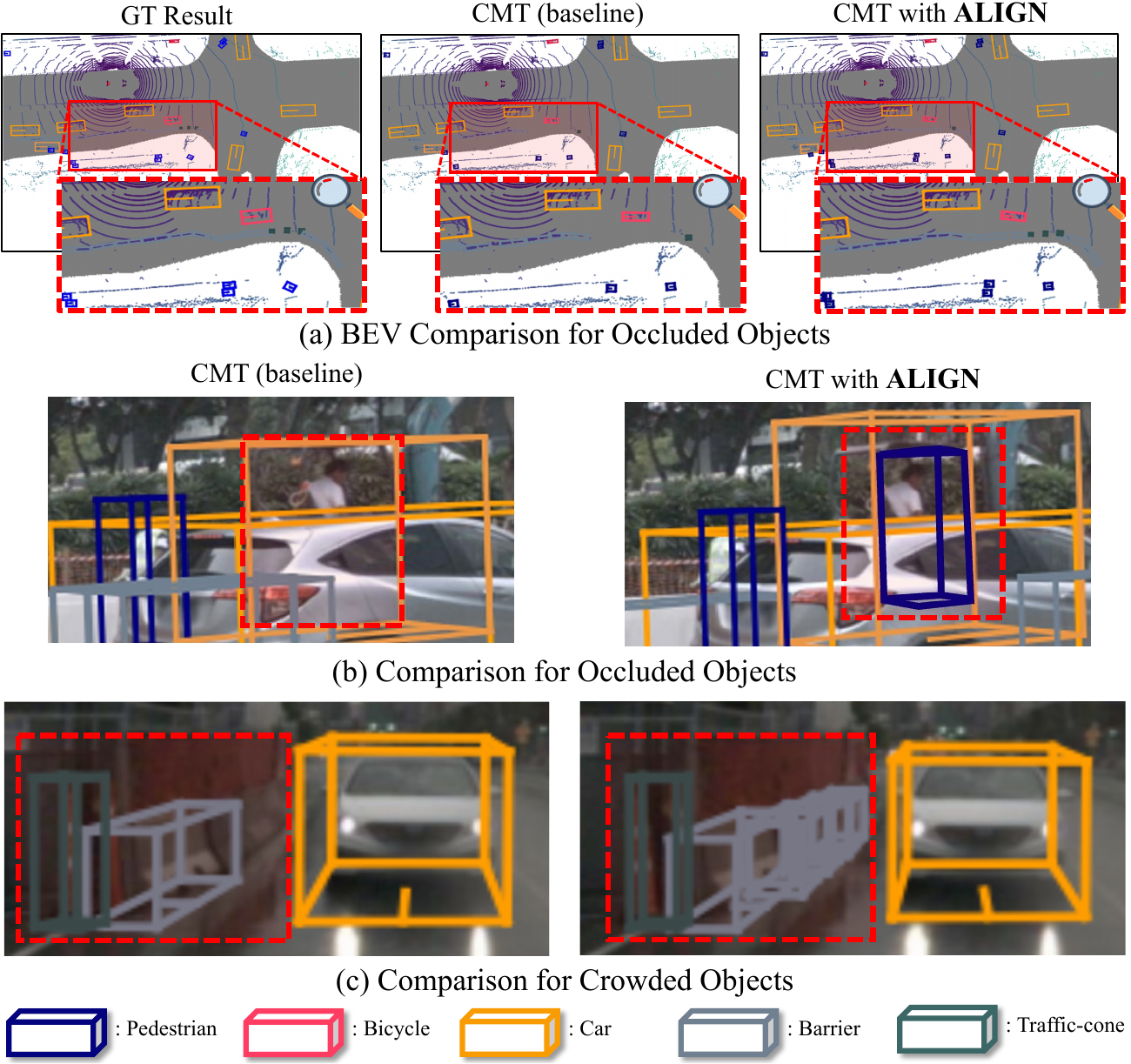}
  \vspace{-0.5em}
  \caption{
\textbf{Qualitative results on occluded and crowded scenes}
}
  \vspace{-1.5 em}
  \label{fig:qualitative_supple}
\end{figure*}

Figure~\ref{fig:qualitative_supple} shows qualitative comparisons between the ground truth, the CMT baseline, and CMT with ALIGN on challenging scenes with heavy occlusion and crowding.

In (a), a BEV-level comparison illustrates that while nearby vehicles and bicycles are detected similarly across models, the baseline often misses or mislocalizes partially occluded or distant objects such as barriers and pedestrians. In contrast, ALIGN more reliably recovers these occluded instances and produces more accurate bounding boxes in the highlighted regions.
In (b), image-view results further show that CMT fails to localize pedestrians partially hidden behind vehicles or street furniture, whereas ALIGN successfully detects these occluded pedestrians.
In (c), crowded scenes with overlapping objects remain challenging for the baseline, which often produces incomplete or fragmented detections. ALIGN, however, yields more stable and coherent predictions in such dense environments.

Overall, these results demonstrate that ALIGN substantially improves detection robustness under occlusion and crowding through geometry-aware query initialization. Nevertheless, extremely small and thin objects such as distant traffic cones remain challenging due to limited visual and geometric cues, leaving room for future improvement.

\clearpage

\clearpage

\bigskip



\bibliographystyle{named}
\bibliography{ijcai26}

@inproceedings{van2024multimodal,
  title={Multimodal object query initialization for 3d object detection},
  author={Van Geerenstein, Mathijs R and Ruppel, Felicia and Dietmayer, Klaus and Gavrila, Dariu M},
  booktitle={2024 IEEE International Conference on Robotics and Automation (ICRA)},
  pages={12484--12491},
  year={2024},
  organization={IEEE}
}

@inproceedings{bai_transfusion_2022,
	address = {New Orleans, LA, USA},
	title = {{TransFusion}: {Robust} {LiDAR}-{Camera} {Fusion} for {3D} {Object} {Detection} with {Transformers}},
	isbn = {978-1-66546-946-3},
	shorttitle = {{TransFusion}},
	url = {https://ieeexplore.ieee.org/document/9879824/},
	doi = {10.1109/CVPR52688.2022.00116},
	language = {en},
	urldate = {2022-11-23},
	booktitle = {2022 {IEEE}/{CVF} {Conference} on {Computer} {Vision} and {Pattern} {Recognition} ({CVPR})},
	publisher = {IEEE},
	author = {Bai, Xuyang and Hu, Zeyu and Zhu, Xinge and Huang, Qingqiu and Chen, Yilun and Fu, Hangbo and Tai, Chiew-Lan},
	month = jun,
	year = {2022},
	pages = {1080--1089},
}

@inproceedings{
zhu_deformable_2021,
title={{Deformable DETR: Deformable Transformers for End-to-End Object Detection}},
author={Xizhou Zhu and Weijie Su and Lewei Lu and Bin Li and Xiaogang Wang and Jifeng Dai},
booktitle={International Conference on Learning Representations (ICLR)},
year={2021},
url={https://openreview.net/forum?id=gZ9hCDWe6ke}
}

@misc{chen_futr3d_2022,
	title = {{FUTR3D}: {A} {Unified} {Sensor} {Fusion} {Framework} for {3D} {Detection}},
	shorttitle = {{FUTR3D}},
	url = {http://arxiv.org/abs/2203.10642},
	language = {en},
	urldate = {2022-12-05},
	publisher = {arXiv},
	author = {Chen, Xuanyao and Zhang, Tianyuan and Wang, Yue and Wang, Yilun and Zhao, Hang},
	month = mar,
	year = {2022},
	note = {Issue: arXiv:2203.10642
arXiv:2203.10642 [cs]},
}

@misc{liu_bevfusion_2022_arxiv,
	title = {{BEVFusion}: {Multi}-{Task} {Multi}-{Sensor} {Fusion} with {Unified} {Bird}'s-{Eye} {View} {Representation}},
	shorttitle = {{BEVFusion}},
	url = {http://arxiv.org/abs/2205.13542},
	language = {en},
	urldate = {2022-12-05},
	publisher = {arXiv},
	author = {Liu, Zhijian and Tang, Haotian and Amini, Alexander and Yang, Xinyu and Mao, Huizi and Rus, Daniela and Han, Song},
	month = jun,
	year = {2022},
	note = {Issue: arXiv:2205.13542
arXiv:2205.13542 [cs]},
}

@inproceedings{chen2023focalformer3d,
  title={Focalformer3d: focusing on hard instance for 3d object detection},
  author={Chen, Yilun and Yu, Zhiding and Chen, Yukang and Lan, Shiyi and Anandkumar, Anima and Jia, Jiaya and Alvarez, Jose M},
  booktitle={Proceedings of the IEEE/CVF International Conference on Computer Vision},
  pages={8394--8405},
  year={2023}
}

@misc{li_unifying_2022,
	title = {Unifying {Voxel}-based {Representation} with {Transformer} for {3D} {Object} {Detection}},
	url = {http://arxiv.org/abs/2206.00630},
	language = {en},
	urldate = {2022-12-15},
	publisher = {arXiv},
	author = {Li, Yanwei and Chen, Yilun and Qi, Xiaojuan and Li, Zeming and Sun, Jian and Jia, Jiaya},
	month = oct,
	year = {2022},
	note = {Issue: arXiv:2206.00630
arXiv:2206.00630 [cs]},
}

@incollection{liang_deep_2018,
	address = {Cham},
	title = {Deep {Continuous} {Fusion} for {Multi}-sensor {3D} {Object} {Detection}},
	volume = {11220},
	isbn = {978-3-030-01269-4 978-3-030-01270-0},
	url = {https://link.springer.com/10.1007/978-3-030-01270-0_39},
	language = {en},
	urldate = {2022-12-20},
	booktitle = {Computer {Vision} {\textendash} {ECCV} 2018},
	publisher = {Springer International Publishing},
	author = {Liang, Ming and Yang, Bin and Wang, Shenlong and Urtasun, Raquel},
	year = {2018},
	doi = {10.1007/978-3-030-01270-0_39},
	pages = {663--678},
}

@misc{liu_petr_2022,
	title = {{PETR}: {Position} {Embedding} {Transformation} for {Multi}-{View} {3D} {Object} {Detection}},
	shorttitle = {{PETR}},
	url = {http://arxiv.org/abs/2203.05625},
	language = {en},
	urldate = {2022-12-21},
	publisher = {arXiv},
	author = {Liu, Yingfei and Wang, Tiancai and Zhang, Xiangyu and Sun, Jian},
	month = jul,
	year = {2022},
	note = {Issue: arXiv:2203.05625
arXiv:2203.05625 [cs]},
}

@inproceedings{wang2022detr3d,
  title={Detr3d: 3d object detection from multi-view images via 3d-to-2d queries},
  author={Wang, Yue and Guizilini, Vitor Campagnolo and Zhang, Tianyuan and Wang, Yilun and Zhao, Hang and Solomon, Justin},
  booktitle={Conference on Robot Learning},
  pages={180--191},
  year={2022},
  organization={PMLR}
}

@inproceedings{caesar_nuscenes_2020,
	address = {Seattle, WA, USA},
	title = {{nuScenes}: {A} {Multimodal} {Dataset} for {Autonomous} {Driving}},
	isbn = {978-1-72817-168-5},
	shorttitle = {{nuScenes}},
	url = {https://ieeexplore.ieee.org/document/9156412/},
	doi = {10.1109/CVPR42600.2020.01164},
	language = {en},
	urldate = {2022-12-23},
	booktitle = {2020 {IEEE}/{CVF} {Conference} on {Computer} {Vision} and {Pattern} {Recognition} ({CVPR})},
	publisher = {IEEE},
	author = {Caesar, Holger and Bankiti, Varun and Lang, Alex H. and Vora, Sourabh and Liong, Venice Erin and Xu, Qiang and Krishnan, Anush and Pan, Yu and Baldan, Giancarlo and Beijbom, Oscar},
	month = jun,
	year = {2020},
	pages = {11618--11628},
}

@inproceedings{he2022destr,
  title={Destr: Object detection with split transformer},
  author={He, Liqiang and Todorovic, Sinisa},
  booktitle={Proceedings of the IEEE/CVF conference on computer vision and pattern recognition},
  pages={9377--9386},
  year={2022}
}

@misc{yan_cross_2023,
	title = {Cross {Modal} {Transformer} via {Coordinates} {Encoding} for {3D} {Object} {Dectection}},
	url = {http://arxiv.org/abs/2301.01283},
	language = {en},
	urldate = {2023-02-20},
	publisher = {arXiv},
	author = {Yan, Junjie and Liu, Yingfei and Sun, Jianjian and Jia, Fan and Li, Shuailin and Wang, Tiancai and Zhang, Xiangyu},
	month = jan,
	year = {2023},
	note = {Issue: arXiv:2301.01283
arXiv:2301.01283 [cs]},
}

@article{zhou_centerformer_2022,
	title = {{CenterFormer}: {Center}-{Based} {Transformer} for {3D} {Object} {Detection}},
	volume = {13698},
	shorttitle = {{CenterFormer}},
	url = {https://link.springer.com/10.1007/978-3-031-19839-7_29},
	language = {en},
	urldate = {2023-03-07},
	journal = {Computer Vision {\textendash} ECCV 2022},
	author = {Zhou, Zixiang and Zhao, Xiangchen and Wang, Yu and Wang, Panqu and Foroosh, Hassan},
	year = {2022},
	doi = {10.1007/978-3-031-19839-7_29},
	pages = {496--513},
}

@misc{smith_cyclical_2017,
	title = {Cyclical {Learning} {Rates} for {Training} {Neural} {Networks}},
	url = {http://arxiv.org/abs/1506.01186},
	language = {en},
	urldate = {2023-03-29},
	publisher = {arXiv},
	author = {Smith, Leslie N.},
	month = apr,
	year = {2017},
	note = {Issue: arXiv:1506.01186
arXiv:1506.01186 [cs]},
}

@misc{zhu_class-balanced_2019,
	title = {Class-balanced {Grouping} and {Sampling} for {Point} {Cloud} {3D} {Object} {Detection}},
	url = {http://arxiv.org/abs/1908.09492},
	language = {en},
	urldate = {2023-04-06},
	publisher = {arXiv},
	author = {Zhu, Benjin and Jiang, Zhengkai and Zhou, Xiangxin and Li, Zeming and Yu, Gang},
	month = aug,
	year = {2019},
	note = {Issue: arXiv:1908.09492
arXiv:1908.09492 [cs]},
}

@inproceedings{carion2020end,
  title={End-to-end object detection with transformers},
  author={Carion, Nicolas and Massa, Francisco and Synnaeve, Gabriel and Usunier, Nicolas and Kirillov, Alexander and Zagoruyko, Sergey},
  booktitle={European conference on computer vision},
  pages={213--229},
  year={2020},
  organization={Springer}
}

@article{kuhn_hungarian_1955,
	title = {The {Hungarian} method for the assignment problem},
	volume = {2},
	issn = {1931-9193},
	url = {https://onlinelibrary.wiley.com/doi/abs/10.1002/nav.3800020109},
	doi = {10.1002/nav.3800020109},
	language = {en},
	number = {1-2},
	urldate = {2023-06-17},
	journal = {Naval Research Logistics Quarterly},
	author = {Kuhn, H. W.},
	year = {1955},
	pages = {83--97},
}

@inproceedings{lin_focal_2017,
	title = {Focal {Loss} for {Dense} {Object} {Detection}},
	url = {https://openaccess.thecvf.com/content_iccv_2017/html/Lin_Focal_Loss_for_ICCV_2017_paper.html},
	urldate = {2023-06-18},
	booktitle = {Proceedings of the {IEEE} {International} {Conference} on {Computer} {Vision}},
	author = {Lin, Tsung-Yi and Goyal, Priya and Girshick, Ross and He, Kaiming and Dollar, Piotr},
	year = {2017},
	pages = {2980--2988},
}

@inproceedings{chen_futr3d_2023,
	title = {{FUTR3D}: {A} {Unified} {Sensor} {Fusion} {Framework} for {3D} {Detection}},
	shorttitle = {{FUTR3D}},
	url = {https://openaccess.thecvf.com/content/CVPR2023W/WAD/html/Chen_FUTR3D_A_Unified_Sensor_Fusion_Framework_for_3D_Detection_CVPRW_2023_paper.html},
	language = {en},
	urldate = {2023-06-18},
	booktitle = {Proceedings of the {IEEE}/{CVF} {Conference} on {Computer} {Vision} and {Pattern} {Recognition} {(CVPR) Workshops}},
	author = {Chen, Xuanyao and Zhang, Tianyuan and Wang, Yue and Wang, Yilun and Zhao, Hang},
	year = {2023},
	pages = {172--181},
}

@misc{yan2023crossmodaltransformerfast,
      title={Cross Modal Transformer: Towards Fast and Robust 3D Object Detection}, 
      author={Junjie Yan and Yingfei Liu and Jianjian Sun and Fan Jia and Shuailin Li and Tiancai Wang and Xiangyu Zhang},
      year={2023},
      eprint={2301.01283},
      archivePrefix={arXiv},
      primaryClass={cs.CV},
      url={https://arxiv.org/abs/2301.01283}, 
}

@inproceedings{he2017mask,
  title={Mask r-cnn},
  author={He, Kaiming and Gkioxari, Georgia and Doll{\'a}r, Piotr and Girshick, Ross},
  booktitle={Proceedings of the IEEE international conference on computer vision},
  pages={2961--2969},
  year={2017}
}

@misc{loshchilov2019decoupledweightdecayregularization,
      title={Decoupled Weight Decay Regularization}, 
      author={Ilya Loshchilov and Frank Hutter},
      year={2019},
      eprint={1711.05101},
      archivePrefix={arXiv},
      primaryClass={cs.LG},
      url={https://arxiv.org/abs/1711.05101}, 
}

@inproceedings{xie2023sparsefusion,
  title={Sparsefusion: Fusing multi-modal sparse representations for multi-sensor 3d object detection},
  author={Xie, Yichen and Xu, Chenfeng and Rakotosaona, Marie-Julie and Rim, Patrick and Tombari, Federico and Keutzer, Kurt and Tomizuka, Masayoshi and Zhan, Wei},
  booktitle={Proceedings of the IEEE/CVF International Conference on Computer Vision},
  pages={17591--17602},
  year={2023}
}

@inproceedings{chen2022deformable,
  title={Deformable feature aggregation for dynamic multi-modal 3D object detection},
  author={Chen, Zehui and Li, Zhenyu and Zhang, Shiquan and Fang, Liangji and Jiang, Qinhong and Zhao, Feng},
  booktitle={European conference on computer vision},
  pages={628--644},
  year={2022},
  organization={Springer}
}

@article{schubert2017dbscan,
  title={DBSCAN revisited, revisited: why and how you should (still) use DBSCAN},
  author={Schubert, Erich and Sander, J{\"o}rg and Ester, Martin and Kriegel, Hans Peter and Xu, Xiaowei},
  journal={ACM Transactions on Database Systems (TODS)},
  volume={42},
  number={3},
  pages={1--21},
  year={2017},
  publisher={Acm New York, NY, USA}
}

@misc{wang2023objectquerylifting2d,
      title={Object as Query: Lifting any 2D Object Detector to 3D Detection}, 
      author={Zitian Wang and Zehao Huang and Jiahui Fu and Naiyan Wang and Si Liu},
      year={2023},
      eprint={2301.02364},
      archivePrefix={arXiv},
      primaryClass={cs.CV},
      url={https://arxiv.org/abs/2301.02364}, 
}

@inproceedings{gao2022adamixer,
  title={Adamixer: A fast-converging query-based object detector},
  author={Gao, Ziteng and Wang, Limin and Han, Bing and Guo, Sheng},
  booktitle={Proceedings of the IEEE/CVF Conference on Computer Vision and Pattern Recognition},
  pages={5364--5373},
  year={2022}
}

@inproceedings{zhao2024detrs,
  title={Detrs beat yolos on real-time object detection},
  author={Zhao, Yian and Lv, Wenyu and Xu, Shangliang and Wei, Jinman and Wang, Guanzhong and Dang, Qingqing and Liu, Yi and Chen, Jie},
  booktitle={Proceedings of the IEEE/CVF Conference on Computer Vision and Pattern Recognition},
  pages={16965--16974},
  year={2024}
}

@inproceedings{ji2024enhancing,
  title={Enhancing 3d object detection with 2d detection-guided query anchors},
  author={Ji, Haoxuanye and Liang, Pengpeng and Cheng, Erkang},
  booktitle={Proceedings of the IEEE/CVF Conference on Computer Vision and Pattern Recognition},
  pages={21178--21187},
  year={2024}
}

@article{paszke2019pytorch,
  title={Pytorch: An imperative style, high-performance deep learning library},
  author={Paszke, Adam and Gross, Sam and Massa, Francisco and Lerer, Adam and Bradbury, James and Chanan, Gregory and Killeen, Trevor and Lin, Zeming and Gimelshein, Natalia and Antiga, Luca and others},
  journal={Advances in neural information processing systems},
  volume={32},
  year={2019}
}

@article{raschka2020machine,
  title={Machine Learning in Python: Main developments and technology trends in data science, machine learning, and artificial intelligence},
  author={Raschka, Sebastian and Patterson, Joshua and Nolet, Corey},
  journal={arXiv preprint arXiv:2002.04803},
  year={2020}
}

@misc{DLPack,
  author = {{DLPack Contributors}},
  title = {{DLPack: Open In-Memory Tensor Structure}},
  howpublished = {\url{https://github.com/dmlc/dlpack}},
  year = {2020}
}

@article{yang2024reparameterizable,
  title={Reparameterizable dual-resolution network for real-time semantic segmentation},
  author={Yang, Guoyu and Wang, Yuan and Shi, Daming},
  journal={arXiv preprint arXiv:2406.12496},
  year={2024}
}

@article{cai2019cascade,
  title={Cascade R-CNN: High quality object detection and instance segmentation},
  author={Cai, Zhaowei and Vasconcelos, Nuno},
  journal={IEEE transactions on pattern analysis and machine intelligence},
  volume={43},
  number={5},
  pages={1483--1498},
  year={2019},
  publisher={IEEE}
}

@inproceedings{bolya2019yolact,
  title={Yolact: Real-time instance segmentation},
  author={Bolya, Daniel and Zhou, Chong and Xiao, Fanyi and Lee, Yong Jae},
  booktitle={Proceedings of the IEEE/CVF international conference on computer vision},
  pages={9157--9166},
  year={2019}
}

@article{wang2020solov2,
  title={Solov2: Dynamic and fast instance segmentation},
  author={Wang, Xinlong and Zhang, Rufeng and Kong, Tao and Li, Lei and Shen, Chunhua},
  journal={Advances in Neural information processing systems},
  volume={33},
  pages={17721--17732},
  year={2020}
}

@inproceedings{zhang2024sparselif,
  title={SparseLIF: High-performance sparse LiDAR-camera fusion for 3D object detection},
  author={Zhang, Hongcheng and Liang, Liu and Zeng, Pengxin and Song, Xiao and Wang, Zhe},
  booktitle={European Conference on Computer Vision},
  pages={109--128},
  year={2024},
  organization={Springer}
}

@article{gong2025roadside,
  title={Roadside LiDAR-camera fusion detection based on spatiotemporal calibration},
  author={Gong, Bowen and Wang, Yimeng and Lin, Ciyun and Liu, Hongchao},
  journal={IEEE Sensors Journal},
  year={2025},
  publisher={IEEE}
}

@inproceedings{huang2024detecting,
  title={Detecting as labeling: Rethinking lidar-camera fusion in 3d object detection},
  author={Huang, Junjie and Ye, Yun and Liang, Zhujin and Shan, Yi and Du, Dalong},
  booktitle={European Conference on Computer Vision},
  pages={439--455},
  year={2024},
  organization={Springer}
}

\end{document}